\documentclass[10pt, conference, compsocconf]{IEEEtran}
\ifCLASSINFOpdf

\else

\fi

\usepackage{amssymb}
\usepackage{graphicx}
\usepackage{times}
\usepackage{epsfig}
\usepackage{amsmath}
\usepackage{subfigure}
\usepackage{amssymb}
\usepackage{algorithm}
\usepackage{algorithmic}
\usepackage{epstopdf}
\usepackage{mathtools}
\DeclareMathOperator*{\argmin}{argmin}

\begin{document}
\title{An Approach for Noise Removal on Depth Images}

\author{\IEEEauthorblockN{Rashi Chaudhary}
\IEEEauthorblockA{Fakir Mohan University}

\and
\IEEEauthorblockN{Himanshu Dasgupta}
\IEEEauthorblockA{Indus International University}
\vspace{-6ex}}
\maketitle
\vspace{-55cm}

\begin{abstract}
Image based rendering is a fundamental problem in computer vision and graphics. Modern techniques often rely on depth image for the 3D construction. However for most of the existing depth cameras, the large and unpredictable noises can be problematic, which can cause noticeable artifacts in the rendered results. In this paper, we proposed an efficacious method for depth image noise removal that can be applied for most RGB-D systems. The proposed solution will benefit many subsequent vision problems such as 3D reconstruction, novel view rendering, object recognition. Our experimental results demonstrate the efficacy and accuracy.
\end{abstract}

\IEEEpeerreviewmaketitle

\section{Introduction}

Nowadays, depth sensors are getting increasingly popular that receive considerable attentions from researchers \cite{Khoshelham12}. The low-cost and realtime time features of depth sensor facilitate many vision tasks, such as image rendering, 3D reconstruction, image localization \cite{Shenicme13}. However, among most of existing systems, the acquired depth image often suffers excessive noises that has degraded its performance and resulted in inaccurate estimation \cite{apsipa12}.

Many researcher have proposed different algorithms to resolve the above issue. For example, in \cite{Fleishman03}, the authors try to use bilateral filter based method to remove the noises while preserve the edges. Similar techniques can also be found in \cite{Fu13}. However, this types of methods fail for depth images of complex scenes as it wipe out many weak edges, which makes the resulting image less faithful to the original settings. Some other researchers casted the denoising process to a image in-painting problem by adopting exemplar-copy scheme \cite{Alvarez02}\cite{Khoshabeh11}. In recent papers, probabilistic framework based methods were introduced by considering depth denoising problem as a labeling process that cluster the image into multiple regions \cite{Shentip13}\cite{Shencvpr13}. This type of methods demonstrate favorable results. But it is expensive that applicable for realtime applications.

\section{Our Solution}
Our approach for the depth denoising have three main phases involved. First, we use our edge detector to identify salient edges across the image, i.e. we apply the traditional Canny edge to extract all the possible distinct edges from an image. Second, we apply jointly bilateral filter to the image smoothen those regions with concinnous texture values while skips processing the parts with distinct structures \cite{Xiao12}. By this step, we can cluster the whole image by using the extracted structures by using those salient edges \cite{Achanta09}.

\begin{equation}
I_t = \frac{1}{k_p}\cdot\sum_{q\in \Omega}I_qf(||p - q||)\cdot g(||I_p - I_q||)
\end{equation}
where $p$ and $q$ represent the center location on image $I$ for the Gaussian Kernels $f$ and $g$. $k_p$ is the normalizing factor; $\Omega$ is the spatial range. After this step, we use the exemplar based scheme to find an optimal patch from those available depth region to the target region. During the filling procedure, only patches the same region enclosed by extracted structures are used for the depth inference. Here we adopt the strategy as Criminisi proposed on isophote-driven sampling process \cite{Criminisi03}.

\begin{equation}
C(p)=\frac{\sum_{q\in\Phi_p\cap\Omega}C(q)}{|\Phi_p|} \mbox{, } D(p) = \frac{\nabla I_p^\bot \cdot n_p}{\alpha}
\end{equation}
where the metrics $C(p)$ and $D(p)$ are the confidence term and data term for the priority patch definition $P(p) = C(p)\cdot D(p)$. The difference from our method from the one \cite{Criminisi03} is that only patches are clustered from the same region will be considered for the target region for in-painting. For each patch, a priority metric is assigned, which determines the order for the target region to fill. Similar to \cite{Criminisi03}, we search in the source region for the potential textures to fill the target region. As only search patches that are clusterd from the same ground by the edges from step 1. Our $\Psi_{\hat{p}}$ is formally defined as:

\begin{equation}
\psi_{\hat{q}} = \argmin_{\psi_q \in \Phi} d(\Psi_{\hat{p}}, \Psi_q) \mbox{, where } p, q \in R_i | i \in\{ 1, 2, ..., n\}
\end{equation}

Solving this optimization problem yields satisfactory depth map. Also, to speed up the performance, we apply histogram-based clustering to the image before the edge extraction which make the relisted structure with less distortion.

\section{The Results}
Our proposed methods have been tested on two public dataset: Tsukuba Stereo database, which was collected by the Tsukuba University, and the frame sequence of ``Ballet'' captured from Microsoft research. Since the ground truth are provided, it allows us to evaluate our results on the accuracy of noise removal and computational cost. All the experiments were carried out on a PC with Intel(R) Core (TM) CPU E5-2620 v2@ 3.50GHz with 24.0GB RAM. First, we applied various patch sizes for the in-paiting as Table \ref{table:1} shows. In the table, it demonstrated 8 samples from the more than 1800 images, has leads to significant improvement in PSNR (Peak Signal-to-noise ratio). As can be seen from the table, our algorithm has high accuracy on different patch sizes. Furthermore, according to our experiments, our algorithm has stable performance for different samples and various sizes of patches.

\begin{table}[h!]
\begin{tabular}{ |p{1.7cm}||p{1.7cm}|p{1.7cm}|p{1.7cm}|  }
 \hline
 \multicolumn{4}{|c|}{Country List} \\
 \hline
 Image ID & Patch size $5 \times 5$ pixels & Patch size $12 \times 12$ pixels & Patch size $20 \times 20$ pixels\\
 \hline
Sample 206   & 5.262    & 5.231 &  5.520 \\
Sample 12 &   5.712  & 5.801   & 5.725\\
 Sample 1152 & 5.692 & 5.820 &  6.234\\
 Sample 99    & 5.291 & 5.013 &  6.102\\
 Sample 328&   5.233 &  5.702 & 5.913\\
 Sample 1602  & 5.234   & 5.133 & 5.913\\
 Sample 5  &  5.203 &5.238 & 6.110\\
 \hline
\end{tabular}
\caption{Improvement on the PSNR by using different patches}
\label{table:1}
\end{table}

\begin{figure}[htb]
\centerline{\epsfig{figure=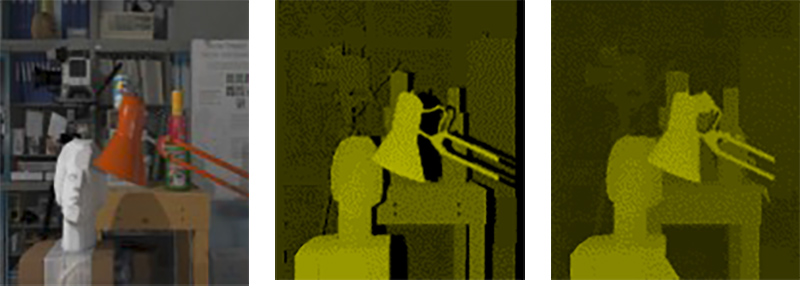, width=9cm}}
\caption{Demo of depth denoising Results: left: original RGB image; middle: original depth image; right: our result image.}\label{fig:evalue1}
\end{figure}

\begin{figure}[htb]
\centerline{\epsfig{figure=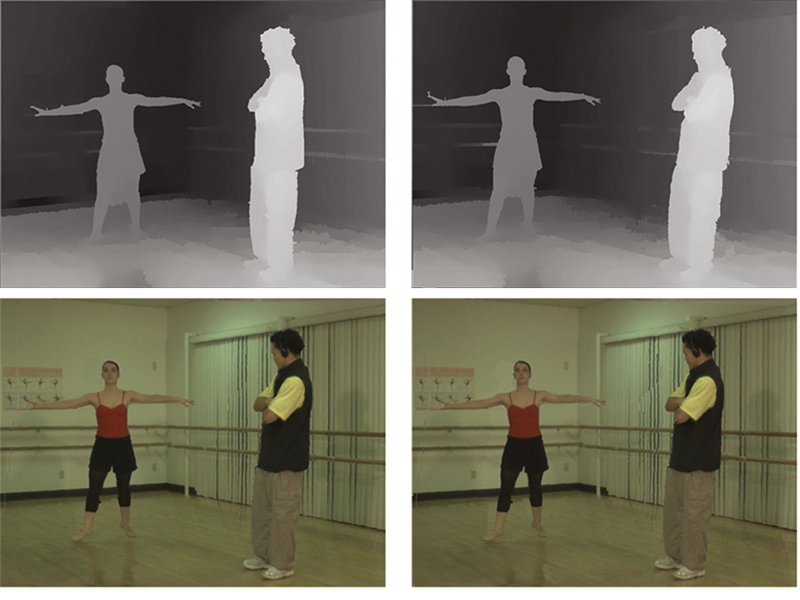, width=9cm}}
\caption{Demo of reconstructed virtual images from the ``Ballet'' sequences: top row: two depth image frames processed by our algorithm; second row: the rendered results by using out refined depth image.}\label{fig:evalue2}
\end{figure}

Figures -\ref{fig:evalue1}, \ref{fig:evalue2} demonstrate the results by running our proposed depth denoising algorithm on the two public data sets. In terms of time performance, we tested on a 200 images with various sizes, it turns out to run reasonably fast, with the average processing time $32$ ms for an image with the size $752\times 520$.

\end{document}